\documentclass{article}

\usepackage{spconf,amsmath,graphicx,hyperref}
\usepackage{cleveref}

\usepackage{algorithm}
\usepackage{algpseudocode}
\usepackage{multirow}
\usepackage[table,xcdraw]{xcolor}
\usepackage{colortbl}
\usepackage{amssymb}
\usepackage{listings}
\usepackage{utfsym}
\usepackage{bbding}
\usepackage{makecell}

\title{From Seeing to Predicting: A Vision-Language Framework for Trajectory Forecasting and Controlled Video Generation}
\name{Fan Yang$^{1,2,3,\star}$, Zhiyang Chen$^{5,\star}$, Yousong Zhu$^{1,3}$, Xin Li$^{2}$, Jinqiao Wang$^{1,2,3,4,\dagger}$ \thanks{
$\dagger$ Corresponding authors: Jinqiao Wang, mail:jqwang@nlpr.ia.ac.cn \\
$\star$ These authors contributed equally to this work.
}
} 
\address{$^{1}$Foundation Model Research Center, Institute of Automation, Chinese Academy of Sciences \\ 
$^{2}$ Peng Cheng Laboratory, Shenzhen, China \\
$^{3}$ School of Artificial Intelligence, University of Chinese Academy of Science, Beijing, China \\
$^{4}$ Wuhan AI Research, Wuhan, China \\
$^{5}$ MAPLE Lab, Westlake University
} 

\begin{document}
%
\maketitle
\begin{abstract}

Current video generation models produce physically inconsistent motion that violates real-world dynamics. We propose TrajVLM-Gen, a two-stage framework for physics-aware image-to-video generation. First, we employ a Vision Language Model to predict coarse-grained motion trajectories that maintain consistency with real-world physics. Second, these trajectories guide video generation through attention-based mechanisms for fine-grained motion refinement. We build a trajectory prediction dataset based on video tracking data with realistic motion patterns. Experiments on UCF-101 and MSR-VTT demonstrate that TrajVLM-Gen outperforms existing methods, achieving competitive FVD scores of 545 on UCF-101 and 539 on MSR-VTT.

\end{abstract}
\section{Introduction}
\label{sec:intro}



Recent video diffusion models (VDMs)~\cite{opensora} have achieved remarkable realism with coherent spatial relationships, rich textures, and realistic lighting effects, demonstrating tremendous potential in content creation and driving research into VDMs as world models. However, these models still struggle to understand and adhere to real-world physics, limiting their reliability in practical applications such as robotic navigation and autonomous driving where physical accuracy is crucial.

Existing approaches have attempted to address this limitation through various strategies. Methods like PixelDance~\cite{zeng2024make} and CameraCtrl~\cite{he2024cameractrl} employ additional visual controls to guide video generation, while Tora~\cite{ToRA} and DragAnything~\cite{wu2024draganything} adopt trajectory-guided diffusion frameworks. However, these methods fundamentally rely on external inputs and lack intrinsic understanding of physical properties such as object dynamics and spatial relationships. Moreover, studies LVD and LVM~\cite{LVD, LVM} leveraging language models for trajectory prediction ignore crucial visual context, limiting their ability to generate physically coherent motions and focusing on superficial motion control rather than fundamental physics understanding.



To bridge the gap between textual descriptions and real-world physical motion, we propose TrajVLM-Gen, a trajectory-guided video generation framework. We leverage the inherent physics "understanding" of vision language models (VLMs) to reason about physical phenomena. For instance, when given a cyclist on a mountain path, VLMs can predict physically plausible trajectories considering gravity and terrain conditions. As shown in~\cref{aaa}, our two-stage approach first employs a vision language model to predict coarse trajectories that conform to real-world dynamics through chain-of-thought reasoning, then leverages these trajectories to guide video generation using attention mechanisms in an OpenSora-based model, achieving fine-grained motion synthesis while maintaining physical consistency.

\begin{figure*}
\centering
\includegraphics[width=0.8\textwidth]{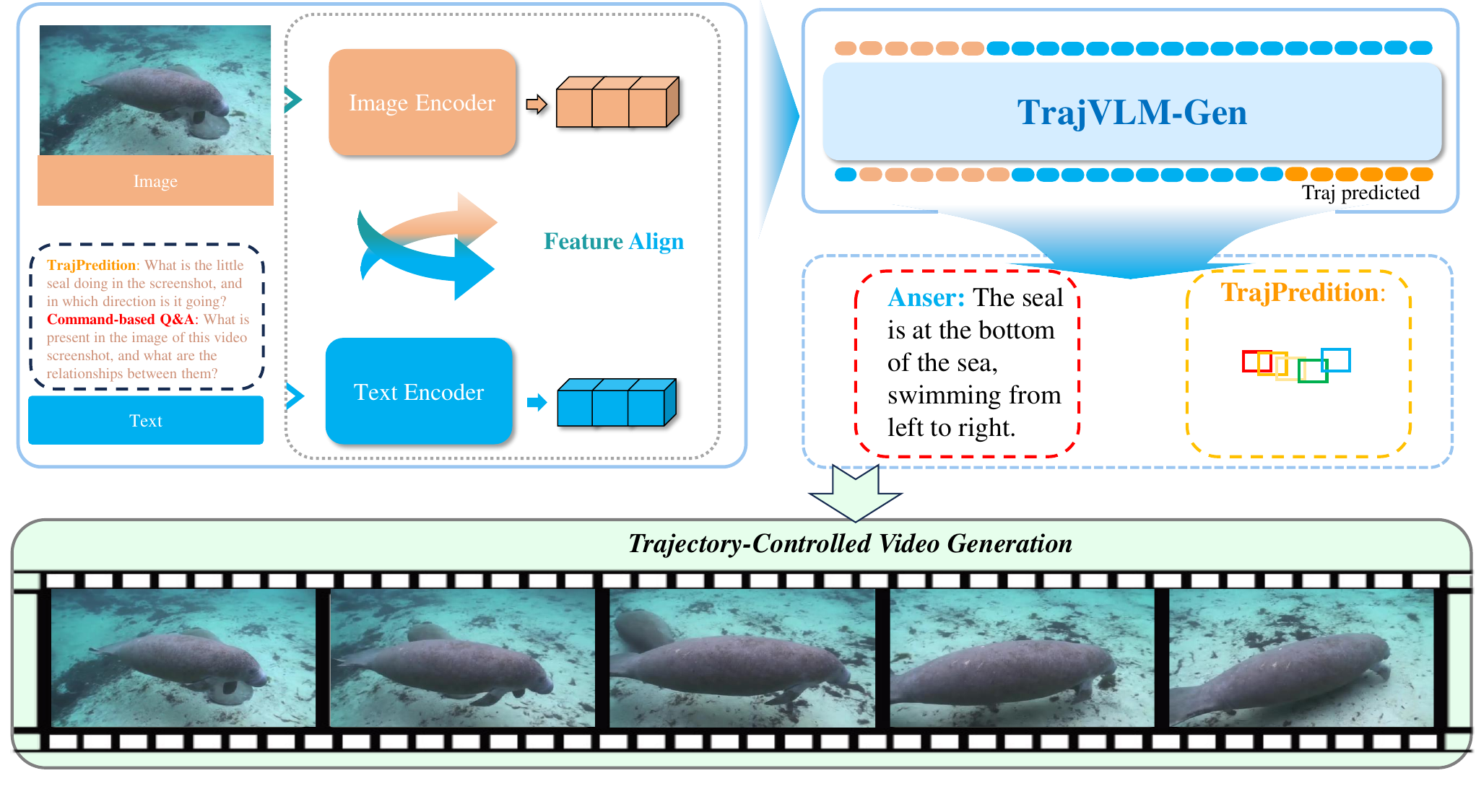}
\label{different}
\vspace{-4mm}
\caption{Our overall framework, TrajVLM-Gen, enables trajectory reasoning and controllable video generation.}
 \label{fig:main}
\end{figure*}


We construct a large-scale dataset with 1.3M image-video-trajectory pairs. Experiments on multiple video generation evaluation tasks including UCF-101 ~\cite{ucf101} and MSR-VTT~\cite{MSR-VTT} demonstrate significant improvements in both trajectory prediction accuracy and video generation quality.

The main contributions include:
\vspace{-2mm}
\begin{enumerate}
\item A vision-language model for trajectory prediction and spatiotemporal modeling;
\vspace{-2mm}
\item A text-guided video generation framework combining trajectory prediction with video synthesis;
\vspace{-2mm}
\item Competitive performance validating trajectory-guided approaches for video generation.
\end{enumerate}

\section{Methodology}
\label{sec:method}

We develop a framework that predicts trajectories from input images and text prompts using Large Vision-Language Models, then generates videos that follow these trajectories. In ~\cref{sec:sec3.1} details the LVLM architecture and trajectory representation, In ~\cref{sec:sec3.2} describes the trajectory-controlled video generation model, and ~\cref{sec:sec3.3} presents dataset construction.

\subsection{Vision-Language Model for Trajectory Prediction}\label{sec:sec3.1}


We achieve object reasoning and trajectory prediction by incorporating numerical coordinates in output text as in ~\cref{fig:main}, following multimodal understanding approaches~\cite{Qwen-vl, Ferret}.



\textbf{Architecture Design.} Our VLM comprises a SigLIP2 vision encoder~\cite{siglip2}, lightweight projection layer, and Qwen2.5-8B language model. The vision encoder extracts visual features $Z_v$ from input images $X_v$, which are mapped to word embedding space $H_v$ and fed into the LLM alongside language tokens $H_{i n s}$ for trajectory prediction.



\textbf{Universal Trajectory Representation.} We develop a unified format to represent temporal trajectories of object bounding boxes. Each bounding box is represented as $p=\left(x_1, y_1, x_2, y_2\right)$ (top-left and bottom-right coordinates),  and trajectories are represented as sequences of bounding boxes $\left[p_1, p_2, \ldots, p_n\right]$.

\textbf{Chain-of-Thought Integration.} To enhance VLM's physical reasoning capabilities, we introduce a chain-of-thought mechanism that formalizes physical phenomenon analysis as a step-by-step reasoning process: (1) analyze video descriptions and identify applicable physical law categories; (2) reason about potential motion patterns of objects in the scene; (3) predict specific changes in bounding boxes over time. The model output format is: "$<$Reason$>$-$\left[p_1, p_2, \ldots, p_n\right]$", where the $<$Reason$>$ field provides the reasoning process. Through this structured process, VLMs can predict future bounding box trajectory changes and generate motion predictions that conform to physical laws.



\subsection{Trajectory-Controlled Video Generation}\label{sec:sec3.2}


We adopt the OpenSora~\cite{opensora} framework for trajectory-guided video generation. This stage leverages VLM-predicted coarse trajectories to guide video diffusion models through self-attention mechanisms, generating physically consistent fine-grained motion. The predicted trajectories are converted to text format $\left[p_1, p_2, \ldots, p_n\right]$ and appended to query text, enabling simultaneous processing of semantic and spatial motion information.

\textbf{Trajectory-Aware Attention Optimization.} 
To ensure generated videos strictly follow predicted trajectories, we use cross-attention to focus on target trajectories. During inference, the model generates cross-attention maps $A_t$ and forms trajectory masks $M_{\text {traj }}$ from coarse-grained trajectory prediction boxes to identify relevant attention regions, guiding the video generation model to enforce precise trajectory constraints. We express this constraint mechanism using an energy function:

\begin{equation}
E\left(A_t\right)=M_{\text {traj }} \odot \operatorname{sign}\left(A_t\right)-\lambda \nabla^2 A_t
\end{equation}

In this function, the term $M_{\text {traj }} \odot \operatorname{sign}\left(A_t\right)$ increases attention values for trajectory-related tokens, ensuring focus remains within intended motion regions. The second term $-\lambda \nabla^2 A_t$ enforces smoothness in attention maps through learnable parameter $\lambda$, preventing spatial abrupt changes. Minimizing this energy function effectively directs the model's attention toward target trajectories, ensuring generated videos accurately follow given motion paths while maintaining overall motion consistency.


\subsection{Video Trajectory Dataset Construction}\label{sec:sec3.3}
We leverage visual tracking datasets to address the scarcity of trajectory prediction training data.


\textbf{Pretraining Data.} We collect multiple public visual tracking datasets, including TNL2K~\cite{TNLK}, LaSOT~\cite{LaSOT-ext}, LaSOT-ext~\cite{LaSOT-ext}, OTB99-Lang~\cite{OTB99-Lang}, and GOT-10K~\cite{GOT-10k}. These datasets provide rich real-world physical motion trajectories with precise target localization. We employ reverse localization to train the model to predict subsequent bounding box positions from the first frame. Each video is segmented into 24-frame clips to maintain motion predictability and appropriate trajectory length. Data preprocessing uses Qwen2.5-VL-72B to remove videos with description errors, low resolution, or tracking interruptions, and applies aesthetic quality filtering (threshold 6.5 using CLIP+MLP scorer~\cite{xu2023unifying}) and optical flow consistency validation (UniMatch model~\cite{xu2023unifying}).

\begin{table}[htbp]
\centering
\caption{Overview of our video trajectory dataset with global description, pretraining, and instruction components.}
\label{tab:datasets}
\scalebox{0.8}{\begin{tabular}{l|l|c|l}
\hline
\textbf{Type} & \textbf{Format} & \textbf{\#Samples} & \textbf{Data Sources} \\ \hline
Global Alignment & Caption & 650K & LLaVA1.5-665K~\cite{llava15} \\ \hline
\multirow{6}{*}{\begin{tabular}[c]{@{}l@{}}Trajectory\\Perception\end{tabular}} 
& \multirow{5}{*}{\begin{tabular}[c]{@{}l@{}}Trajectory\\Pretraining\end{tabular}} 
& \multirow{5}{*}{600K} 
& TNLK2K~\cite{TNLK} \\ \cline{4-4}
& & & LaSOT~\cite{LaSOT-ext} \\ \cline{4-4}
& & & LaSOT-ext~\cite{LaSOT-ext} \\ \cline{4-4}
& & & OTB99-lang~\cite{OTB99-Lang} \\ \cline{4-4}
& & & GOT-10K~\cite{GOT-10k} \\ \cline{2-4}
& \begin{tabular}[c]{@{}l@{}}Trajectory\\Instruction\end{tabular} & 50K & Random sampling \\ \hline
\textbf{Total} & \textbf{--} & \textbf{1.3M} & \textbf{--} \\ \hline
\end{tabular}}
\end{table}

\textbf{Instruction Following Data.} We construct an instruction-following dataset integrating physical reasoning based on pretraining data. Using Gemini2.5-pro to describe video content and assess physical plausibility, we add physical labels based on motion patterns. We analyze bounding box velocity and acceleration to assign physical labels. Trajectories with vertical acceleration are labeled "gravity", those with frequent direction changes or rebounds are labeled "elastic", and scale-varying motion is labeled "perspective camera projection". This creates a physics-aware trajectory prediction dataset supporting natural language guidance for physically plausible trajectory generation. The dataset will be publicly released.


\begin{table*}
\caption{Performance comparison with state-of-the-art methods on popular MLLM benchmarks and general VQA tasks. - Indicates the result is not reported in the original paper.}
\label{tab:nlp}
\centering
\renewcommand{\arraystretch}{1.2}
\scalebox{0.72}{
\begin{tabular}{l|l|l|lllll|llll}
\hline
\rowcolor{white}
Methods      & LLM         & Max Res & POPE & $MME^p$ & $MME^c$ & MMB  & MMMU & VQAv2 & GQA  & SQA  & VisWiz \\ \hline
BLIP-2~\cite{BLIP-2}       & Vicuna-13B  & 224     & 85.3 & 1294                   & -                      & -    & -    & 65.0  & 41.0 & -    & 19.6   \\
InstructBLIP~\cite{InstructBLIP} & Vicuna-7B   & 224     & -    & -                      & -                      & 36.0 & -    & -     & 49.2 & -    & 34.5   \\
Shikra \cite{Shikra}      & Vicuna-13B  & 224     & 83.9 & -                      & -                      & -    & -    & 77.5  & -    & -    & -      \\
Qwen-VL \cite{Qwen-vl}     & Qwen-7B     & 448     & -    & 1488                   & 361                    & 60.6 & -    & 78.2  & 57.5 & 68.2 & 38.9   \\
LLaVA1.5 \cite{llava15}    & Vicuna-13B  & 336     & 85.9 & 1531                   & 295                    & \underline{67.7} & 34.8 & 80.0  & 62.9 & 69.1 & 46.8   \\
InternVL \cite{chen2024internvl}     & Vicuna-13B  & 336     & \underline{87.1} & \textbf{1547}                   & -                      & -    & -    & 80.2  & \underline{63.9} & -    & 54.6   \\
\rowcolor{cyan!20} \textbf{TrajVLM-Gen}  & Qwen2.5-8B  & 224     & 86.8 & 1543 & \underline{298}  & \textbf{68.1} & \underline{35.8} & \underline{80.8}  & 63.1 & \underline{69.5} & \underline{56.8}   \\
\rowcolor{cyan!20} \textbf{TrajVLM-Gen}  & Qwen2.5-8B  & 384     & \textbf{88.1} & \underline{1545} & \textbf{301}  & 67.6 & \textbf{36.3} & \textbf{81.2}  & \textbf{64.3} & \textbf{70.2} & \textbf{57.9}   \\ \hline
\end{tabular}}
\vspace{-2mm}
\end{table*}

\section{Experiments}

We present the experimental setup (~\cref{sec:sec4.1}), compare TrajVLM-Gen with SOTA methods (~\cref{sec:sec4.2}), and verify key components through ablation studies (~\cref{sec:sec4.3}).


\begin{table*}
\caption{Detection-based evaluation of generated videos using LVD shows that our performance outperforms other methods.}
\label{tab:det}
\centering
\scalebox{0.8}{
\begin{tabular}{cccccccc}
\hline
Method          & LLM\_grounded & Numeracy & Attribution & Visibility & Dynamics & Sequential & Average \\ \hline
ModelScope~\cite{ModelScope}      & ×             & 32\%     & 54\%        & 8\%        & 21\%     & 0\%        & 23.0\%  \\
Retrieval-based &  \Checkmark            & 15\%     & 15\%        & 11\%       & 9\%      & 0\%        & 9.7\%   \\
LVD(GPT-3.5) \cite{LVD}    &   \Checkmark           & 52\%     & 79\%        & 64\%       & 37\%     & 2\%        & 46.4\%  \\
LVD(GPT-4) \cite{LVD}      &    \Checkmark          & 41\%     & 64\%        & 55\%       & 51\%     & 38\%       & 49.4\%  \\
\rowcolor{cyan!20} \textbf{TrajVLM-Gen} & \Checkmark & \textbf{89\%} & \textbf{95\%} & \textbf{92\%} & \textbf{95\%} & \textbf{86\%} & \textbf{89.6\%} \\ \hline
\end{tabular}}
\vspace{-2mm}
\end{table*}

\subsection{Experimental Setup}
\label{sec:sec4.1}

\textbf{Training Configuration.} TrajVLM-Gen uses Qwen2.5 framework with 384x384 input resolution and the siglip2-so400m-patch14-384 visual encoder. Training involves pretraining and instruction fine-tuning stages, with the instruction fine-tuning stage using batch size 128 and learning rate 2e-5, taking 1 day. The inference process first predicts trajectories, then performs inference on OpenSora. Inference uses 4 A800 GPUs (~1 hour).






\begin{table}
\caption{Comparison of motion-controllable video generation models on public datasets. TrajVLM-Gen achieves competitive performance on both UCF-101 and MSR-VTT.}
\label{tab:public}
\centering
\scalebox{0.7}{
\begin{tabular}{ccc}
\hline
\multicolumn{1}{l}{Method} & FVD @UCF-101 (↓) & FVD @MSR-VTT (↓) \\ \hline
CogVideo (Chinese)~\cite{CogVideo}  & 751             & -              \\
CogVideo (English)~\cite{CogVideo}  & 702             & 1294           \\
MagicVideo \cite{zhou2022magicvideo}& 699             & 1290           \\
Make-A-Video~\cite{Make-A-Video}    & 367             & -              \\
VideoLDM  \cite{blattmann2023align} & 551             & -              \\ \hline
ModelScope  \cite{ModelScope}       & -               & 550            \\
LVD   \cite{LVD}                    & 828             & 565            \\
\rowcolor{cyan!20} \textbf{TrajVLM-Gen}  & \textbf{545}    & \textbf{539}   \\ \hline
\end{tabular}}
\vspace{-1mm}
\end{table}

\subsection{Comparison to State-of-the-Art Methods}\label{sec:sec4.2}

\textbf{Multimodal Understanding Performance.} As is show in ~\cref{tab:nlp}, TrajVLM-Gen achieves competitive performance across MLLM benchmarks. We obtain 81.2\% on VQAv2, 36.3\% on MMMU, and 88.1\% on POPE, showing consistent improvements over existing methods. These results demonstrate that trajectory understanding enhances multimodal reasoning capabilities.


\begin{table}
\caption{TrajVLM-Gen demonstrates significant performance advantages over state-of-the-art motion-controllable video generation models on public datasets.}
 \label{tab:2}
\scalebox{0.7}{\begin{tabular}{ccc}
\hline
\multicolumn{1}{l}{Method} & FVD @UCF-101(↓) & FVD@MSR-VTT(↓) \\ \hline
TrajVLM-Gen w/o attention mask                        & 632             & 558             \\
\rowcolor{cyan!20} TrajVLM-Gen w  attention mask                 & \textbf{545}             & \textbf{539}           \\ \hline
\end{tabular}}
\vspace{-1mm}
\end{table}



\textbf{Trajectory Generation Accuracy.} As is the ~\cref{tab:det} show TrajVLM-Gen achieves 89.6\% average performance, significantly outperforming LVD(GPT-4) at 49.4\%. Our method shows substantial improvements in Numeracy (89\%) and Sequential reasoning (86\%), validating our trajectory-guided approach.

\begin{figure}[!h]
\centering
\includegraphics[width=0.35\textwidth]{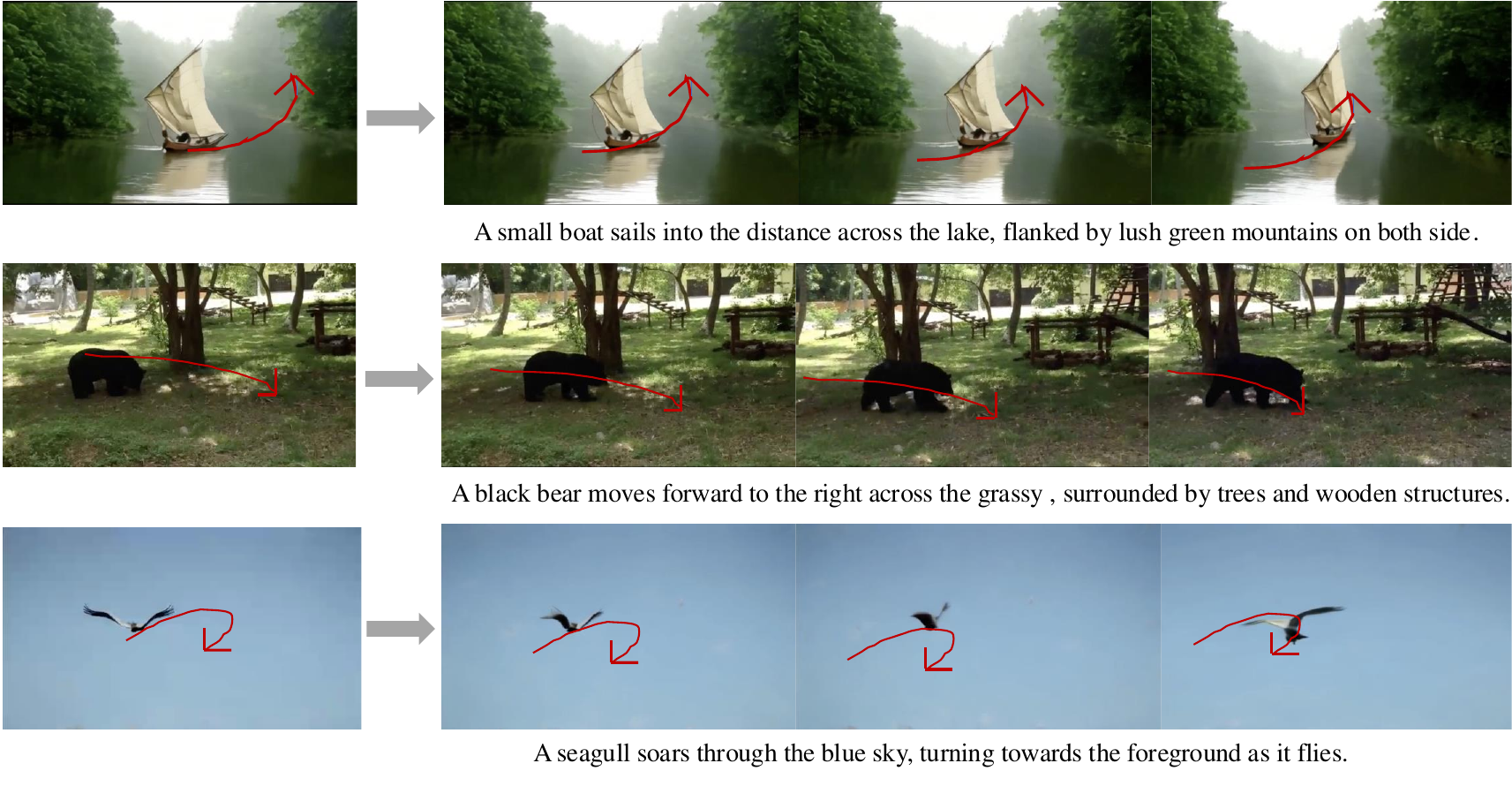}
\vspace{-3mm}
\caption{Based on our predicted trajectories, we can generate videos with reasonable motion that follows physical laws.}
 \label{fig:case}
\end{figure}

\textbf{Video Generation Quality.} As is ~\cref{tab:2} shows our method's performance on public video generation benchmarks. TrajVLM-Gen achieves competitive results on both UCF-101 with FVD of 545 and MSR-VTT with FVD of 539. On MSR-VTT, we outperform recent methods including LVD at 565 and ModelScope at 550. As is show in ~\cref{fig:case} demonstrates our trajectory-guided video generation results, showing how predicted trajectories enable the generation of videos with reasonable motion that follows physical laws. These results validate our approach's effectiveness in producing controllable video content.


\subsection{Ablation Studies}
\label{sec:sec4.3}

\textbf{Effect of Attention Mask Mechanisms.} Table 4 shows trajectory-based cross-attention masks improve accuracy from 67.3\% to 82.1\% (+14.8\%), helping the model focus on relevant trajectory regions despite OpenSora's challenging 3D VAE encoding.

\vspace{-5mm}

\section{Conclusion}

We propose TrajVLM-Gen, a vision-language model that generates controllable videos through trajectory prediction. By representing trajectories as text, our model bridges visual understanding and video generation. Given initial images and text descriptions, TrajVLM-Gen predicts object trajectories and generates corresponding motion videos. We construct a trajectory generation dataset and demonstrate that our approach achieves accurate trajectory prediction while generating high-quality, controllable videos.

------------
\bibliographystyle{IEEEbib}
\bibliography{strings,refs}

\begin{thebibliography}{10}

\bibitem{opensora}
Zangwei Zheng, Xiangyu Peng, Tianji Yang, Chenhui Shen, Shenggui Li, Hongxin Liu, Yukun Zhou, Tianyi Li, and Yang You,
\newblock ``Open-sora: Democratizing efficient video production for all,''
\newblock {\em arXiv preprint arXiv:2412.20404}, 2024.

\bibitem{zeng2024make}
Yan Zeng, Guoqiang Wei, et~al.,
\newblock ``Make pixels dance: High-dynamic video generation,''
\newblock in {\em Proceedings of the IEEE/CVF Conference on Computer Vision and Pattern Recognition}, 2024, pp. 8850--8860.

\bibitem{he2024cameractrl}
Hao He, Yinghao Xu, et~al.,
\newblock ``Cameractrl: Enabling camera control for text-to-video generation,''
\newblock {\em arXiv preprint arXiv:2404.02101}, 2024.

\bibitem{ToRA}
Zhenghao Zhang, Junchao Liao, Menghao Li, Zuozhuo Dai, Bingxue Qiu, Siyu Zhu, Long Qin, and Weizhi Wang,
\newblock ``Tora: Trajectory-oriented diffusion transformer for video generation,''
\newblock {\em arXiv preprint arXiv:2407.21705}, 2024.

\bibitem{wu2024draganything}
Weijia Wu, Zhuang Li, et~al.,
\newblock ``Draganything: Motion control for anything using entity representation,''
\newblock in {\em European Conference on Computer Vision}. Springer, 2024, pp. 331--348.

\bibitem{LVD}
Long Lian, Baifeng Shi, Adam Yala, Trevor Darrell, and Boyi Li,
\newblock ``Llm-grounded video diffusion models,'' 2024.

\bibitem{LVM}
Long Lian, Boyi Li, et~al.,
\newblock ``Llm-grounded diffusion: Enhancing prompt understanding of text-to-image diffusion models with large language models,''
\newblock {\em arXiv preprint arXiv:2305.13655}, 2023.

\bibitem{ucf101}
Khurram Soomro, Amir~Roshan Zamir, et~al.,
\newblock ``Ucf101: A dataset of 101 human actions classes from videos in the wild,'' 2012.

\bibitem{MSR-VTT}
Jun Xu, Tao Mei, et~al.,
\newblock ``Msr-vtt: A large video description dataset for bridging video and language,''
\newblock in {\em Proceedings of the IEEE Conference on Computer Vision and Pattern Recognition (CVPR)}, June 2016.

\bibitem{Qwen-vl}
Jinze Bai, Shuai Bai, et~al.,
\newblock ``Qwen-vl: A versatile vision-language model for understanding, localization, text reading, and beyond,''
\newblock {\em arXiv preprint arXiv:2308.12966}, 2023.

\bibitem{Ferret}
Haoxuan You, Haotian Zhang, Zhe Gan, Xianzhi Du, Bowen Zhang, Zirui Wang, Liangliang Cao, Shih-Fu Chang, and Yinfei Yang,
\newblock ``Ferret: Refer and ground anything anywhere at any granularity,''
\newblock {\em arXiv preprint arXiv:2310.07704}, 2023.

\bibitem{siglip2}
Michael Tschannen, Alexey Gritsenko, et~al.,
\newblock ``Siglip 2: Multilingual vision-language encoders with improved semantic understanding, localization, and dense features,''
\newblock {\em arXiv preprint arXiv:2502.14786}, 2025.

\bibitem{TNLK}
Xiao Wang, Xiujun Shu, et~al.,
\newblock ``Towards more flexible and accurate object tracking with natural language: Algorithms and benchmark,'' 2021.

\bibitem{LaSOT-ext}
Heng Fan, Hexin Bai, et~al.,
\newblock ``Lasot: A high-quality large-scale single object tracking benchmark,'' 2020.

\bibitem{OTB99-Lang}
Yaozong Zheng, Bineng Zhong, et~al.,
\newblock ``Towards unified token learning for vision-language tracking,'' 2023.

\bibitem{GOT-10k}
Lianghua Huang, Xin Zhao, et~al.,
\newblock ``Got-10k: A large high-diversity benchmark for generic object tracking in the wild,''
\newblock {\em IEEE Transactions on Pattern Analysis and Machine Intelligence}, vol. 43, no. 5, pp. 1562–1577, May 2021.

\bibitem{xu2023unifying}
Haofei Xu, Jing Zhang, et~al.,
\newblock ``Unifying flow, stereo and depth estimation,''
\newblock {\em IEEE Transactions on Pattern Analysis and Machine Intelligence}, vol. 45, no. 11, pp. 13941--13958, 2023.

\bibitem{llava15}
Haotian Liu, Chunyuan Li, et~al.,
\newblock ``Improved baselines with visual instruction tuning,'' 2023.

\bibitem{BLIP-2}
Junnan Li, Dongxu Li, et~al.,
\newblock ``Blip-2: Bootstrapping language-image pre-training with frozen image encoders and large language models,'' 2023.

\bibitem{InstructBLIP}
Wenliang Dai, Junnan Li, et~al.,
\newblock ``Instructblip: Towards general-purpose vision-language models with instruction tuning,'' 2023.

\bibitem{Shikra}
Keqin Chen, Zhao Zhang, et~al.,
\newblock ``Shikra: Unleashing multimodal llm's referential dialogue magic,''
\newblock {\em arXiv preprint arXiv:2306.15195}, 2023.

\bibitem{chen2024internvl}
Zhe Chen, Jiannan Wu, et~al.,
\newblock ``Internvl: Scaling up vision foundation models and aligning for generic visual-linguistic tasks,''
\newblock in {\em Proceedings of the IEEE/CVF Conference on Computer Vision and Pattern Recognition}, 2024, pp. 24185--24198.

\bibitem{ModelScope}
Jiuniu Wang, Hangjie Yuan, et~al.,
\newblock ``Modelscope text-to-video technical report,'' 2023.

\bibitem{CogVideo}
Wenyi Hong, Ming Ding, et~al.,
\newblock ``Cogvideo: Large-scale pretraining for text-to-video generation via transformers,'' 2022.

\bibitem{zhou2022magicvideo}
Daquan Zhou, Weimin Wang, et~al.,
\newblock ``Magicvideo: Efficient video generation with latent diffusion models,''
\newblock {\em arXiv preprint arXiv:2211.11018}, 2022.

\bibitem{Make-A-Video}
Uriel Singer, Adam Polyak, et~al.,
\newblock ``Make-a-video: Text-to-video generation without text-video data,'' 2022.

\bibitem{blattmann2023align}
Andreas Blattmann, Robin Rombach, et~al.,
\newblock ``Align your latents: High-resolution video synthesis with latent diffusion models,''
\newblock in {\em Proceedings of the IEEE/CVF conference on computer vision and pattern recognition}, 2023, pp. 22563--22575.

\end{thebibliography}

\end{document}